\documentclass[a4paper,11pt]{article}

\usepackage{cite}
\usepackage{indentfirst}
\usepackage{fourier}
\usepackage{color}
 \usepackage{graphicx}
 \usepackage{subfigure}
 \usepackage{float}
\usepackage{url}
\usepackage[affil-it]{authblk}
\usepackage{amsmath}
\usepackage{wrapfig}

\usepackage{bibspacing}
\usepackage[T1]{fontenc}
\usepackage{times}

\begin{document}
\title{Multi-Person Full Body Pose Estimation}
\author{Haoyi Zhu, Cheng Jie, Shaofei Jiang}
\affil{Shanghai Jiao Tong University, China \authorcr \{zhuhaoyi, jiec\_tech, jiangshaofei\}@sjtu.edu.cn}
\date{}
\maketitle

\thispagestyle{empty}
\vspace{-3em}
\begin{abstract}
Multi-person pose estimation plays an important role in many fields. Although previous works have researched a lot on different parts of human pose estimation, full body pose estimation for multi-person still needs further research. Our work has developed an integrated model through knowledge distillation which can estimate full body poses. Trained based on the AlphaPose system and MSCOCO2017 dataset, our model achieves 51.5 mAP on the validation dataset annotated manually by ourselves. Related resources are available at \url{https://esflfei.github.io/esflfei.gethub.io/website.html}.
\end{abstract}

\section{Introduction}

Multi-person pose estimation has become increasingly popular in computer vision field in recent years. It has many applications such as human-computer interaction, augmented reality, and sports analytics. It can also improve the performance of re-targeting, tracking, and action recognition. 

Previous works on this topic mainly focus on pose estimation of human body or different parts of human, such as head pose estimation and hand pose estimation. However, little research has been conducted on full body pose estimation. OpenPose\cite{Cao} is currently the only system that can estimate multi-person full body pose, which is bottom-up and have to use mutiple networks, and \cite{Hidalgo2019} develops a single network on full body pose estimation based on it, which applies PAF network architecture and multi-task learning to get body part candidates and uses bipartite graph matching to reach the final full body pose.

Our work develops an integrated model to directly estimate multi-person full body pose through a single network based on AlphaPose\cite{Fang2016}, the state-of-the-art multi-person body pose estimation system. The insight of this paper is to train a multi-person full body pose estimation model through knowledge distillation. Our inspiration is from the teacher-student model. The body keypoints groundtruth can be obtained from the annotation of MSCOCO2017 dataset\cite{Lin2014}. Based on them, we can get the predicted keypoints of face, hand and foot. We treat the predicted keypoints as pseudo labels and put all of them together with the body keypoints to get the full body pose label. Thus, we can use it to train a model that can estimate multi-person full body pose.

We train our model on the AlphaPose system and the MSCOCO2017 train dataset. We then annotate full body keypoints on MSCOCO2017 validation dataset by ourselves, where our model reaches a result of 51.5 mAP, 10.0 higher than the latest OpenPose model. Our model performs pretty well on foot and body. When faces or hands are too small or ocludded, the detection accuracy of them will decrease.

\section{Related Work}

There are four parts of research related to our work, including hand pose estimation, face keypoint detection, foot keypoint detection, and body pose estimation. 
\vspace{-2ex}
\paragraph{Hand Pose Estimation}
Due to high cost and challenges in manual annotation of hand keypoint, there does not exist any large hand keypoint dataset. To overcome this problem, Simon et al. \cite{Simon2017} generated a labeled hand keypoint dataset by developing multiview bootstrapping and trained a single view hand keypoint detector. 
\vspace{-1ex}
\paragraph{Face Keypoint Detection}
There are mainly two kinds of approaches to achieve face keypoint detection: regression-based methods and template fitting. Regression Methods rely on Convolutional Neural Networks and often use convolutional heatmap regression, while template fitting usually employ a series of regression functions to fit the original image by creating face templates.
\vspace{-1ex}
\paragraph{Foot Keypoint Detection}
Cao et al. \cite{Cao} developed the first foot dataset based on the COCO dataset. The first detector combining body and foot keypoint was also trained.
\vspace{-1ex}
\paragraph{Body Pose Estimation}
The early way to accomplish body pose estimation is to infer from both local observations and spatial dependencies of body parts. It is divided into two categories: tree-structured graphical-based models and non-tree models. With the development of CNN, the accuracy on body pose estimation grew rapidly and multi-person estimation became possible, mainly containing two ways: top-down and bottom-up.

\section{Approach}

Our aim is to acquire a pseudo label of 133 full body pose keypoints through knowledge distillation, including 17 body keypoints, 6 foot keypoints, 68 face keypoints and 42 hand keypoints (21 per hand), and use them as groundtruth when training. In this paper, we get the pseudo label based on the MSCOCO2017 train dataset. Since the 17 body keypoints have already been labeled in the dataset, we actually only have to obtain the rest and merge all of them together.

\subsection{Data Annotation} 
\begin{figure}[!h]
	\centering
	\subfigure[]{
		\begin{minipage}[t]{0.32\linewidth}
			\centering
			\includegraphics[width=1\textwidth]{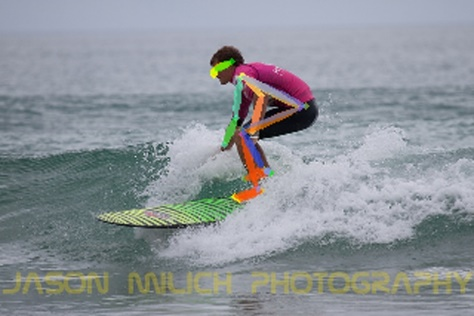}
		\end{minipage}%
	}
	\subfigure[]{
		\begin{minipage}[t]{0.32\linewidth}
			\centering
			\includegraphics[width=1\textwidth,height=0.151\textheight]{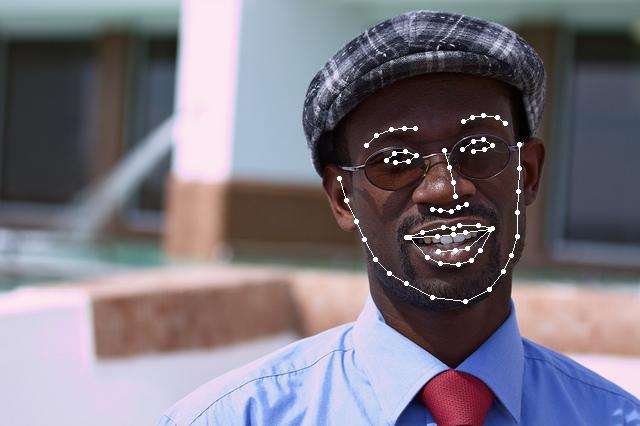}
		\end{minipage}%
	}
	\subfigure[]{
		\begin{minipage}[t]{0.32\linewidth}
			\centering
			\includegraphics[width=1\textwidth,height=0.151\textheight]{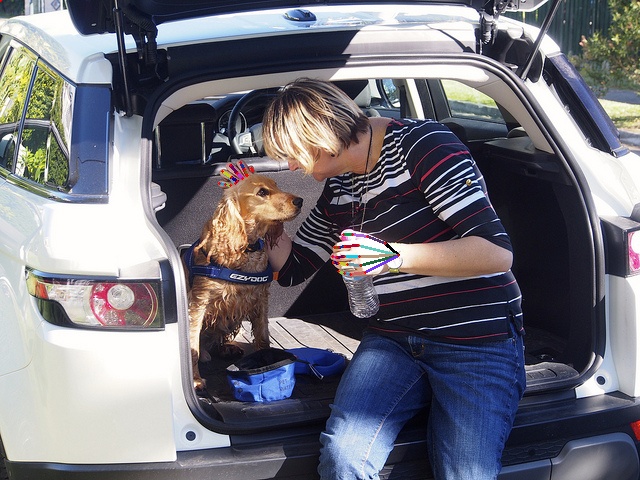}
		\end{minipage}%
	}
	\caption{The visualization results of our data annotation. (a)(b)(c) are examples of the body and foot keypoints, face keypoints and hand keypoints respectively. }
\end{figure}
\vspace{-2ex}
\paragraph{Foot Keypoints}
 Using the existing model based on AlphaPose which detects 17 body keypoints and 6 extended foot keypoints, we can directly obtain the foot keypoints and the visualized connection.
\vspace{-1ex}
\paragraph{Face Keypoints}
Based on the regular architecture of PRNet, we first predict the position of face bounding box using the annotated body parts of nose, eyes and ears. The face bounding box is then cropped and fed into PRNet where the input 2D image is mapped to a corresponding colored UV texture map so that we can predict UV parameters by CNN to do face reconstruction\cite{Feng2018}. Finally, we get the 68 detected face keypoints.
\vspace{-1ex}
\paragraph{Hand Keypoints}
In this section, we extract the hand detection block in OpenPose independently for the output of 21 keypoints per hand. We use annotated body parts to predict the hand box proposals. The proposals then go through a hand estimation model based on Multiview Bootstrapped Training\cite{Simon2017}, which generates geometrically corresponding hand keypoints annotations under an external supervision source of multiple views and uses these annotations to further improve the detector. Finally, the detected hand keypoints are added to the end of keypoints list.
\begin{figure}[H] 
	\centering 
	\begin{minipage}[b]{0.48\textwidth} 
		\centering 
		\includegraphics[width=1\textwidth]{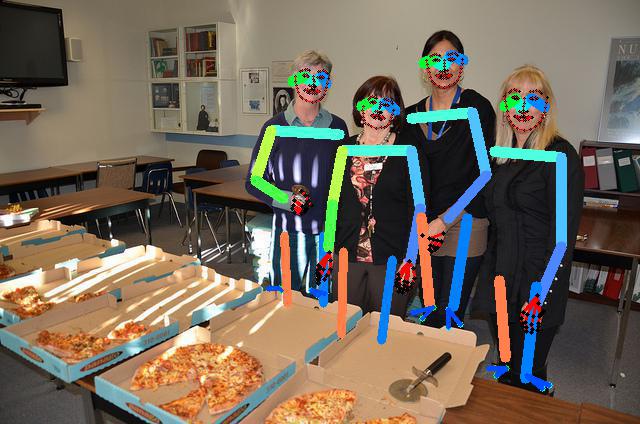} 
	\end{minipage}%
	\hspace{0.02\textwidth}%
	\begin{minipage}[b]{0.48\textwidth} 
		\centering 
		\includegraphics[width=1\textwidth]{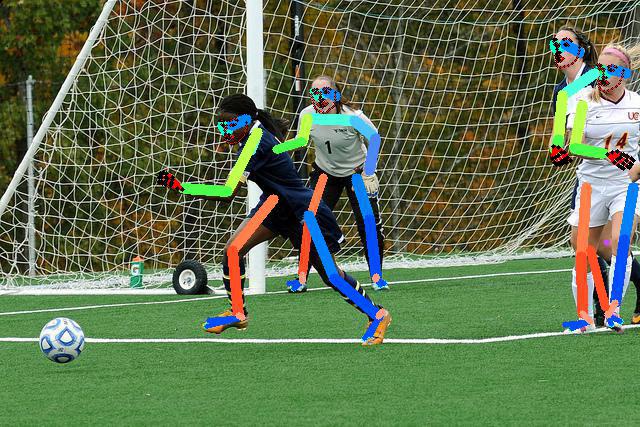} 
	\end{minipage}
	\caption{The final result of full body pose annotation.}
\end{figure}
Now a multi-person full body keypoints annotation containing body, face, hand and foot keypoints is constructed. The merged json can then be utilized as the ground truth for the subsequent training. It can be directly thrown into a single mature pose estimation network (AlphaPose in this paper) to train a full body pose estimation model. 

\subsection{System Framework}
Our work is based on the AlphaPose system which follows the RMPE framework\cite{Fang2016}. It is SPPE-based and follows the two-step framework which contains human detection and pose estimation. The whole framework mainly consists of three blocks: Symmetric Spatial Transformer Network (SSTN), Parametric Pose Non-Maximum-Suppression (NMS) and Pose-Guided Proposals Generator (PGPG). The SSTN attached to the SPPE is used for obtaining an accurate single person region from a rough bounding box. After detecting human proposals, SSTN transforms the proposals to make them more suitable for SPPE, and de-transforms them after SPPE. To improve this step, in the training process, there is an added parallel SPPE branch which back-propagates the center-located pose errors. A parametric pose NMS is then introduced to eliminate redundant pose estimations, which defines a novel pose distance metric with a data-driven optimizer. Finally, the PGPG is employed for data augmentation by learning the output distribution of human detection results.

\subsection{Training}
Our model is trained using seven Nvidia GeForce RTX 2080 Ti graphics cards. Our batch size is set to 8 and the Adam optimizer is used. The input resolution is 256×192 while the output heatmap resolution is set to 64×48. The initial learning rate is set to 0.001 with a learning factor of 0.1. Our model is trained and iterated through 328 epochs to improve performance.

\section{Experiment}

We have manually annotated the full body keypoints on validation dataset of MSCOCO2017\cite{Lin2014}, which includes 5000 images, to evaluate our model. We choose YOLO as our human detector, due to its high efficiency and accuracy. We apply AlphaPose to train our model. Figure 3 shows the visualization result of our model. We then run the OpenPose system, using both the model\_25 (the common model of OpenPose\cite{Cao}) plus face and hand flag and model\_135 (the model trained by [5]), and compare its results with ours, which is shown in \textbf{table 1}. We can see that our mAP reaches 51.5, 10.0 higher than \cite{Hidalgo2019} and 13.3 higher than \cite{Cao}. Among the three methods, our AP and AR under any condition are all the best.

\begin{table}[H]
	\centering
	\begin{tabular}{c|c|cccc|c|cccc}
		\hline
		Methods               & \textbf{mAP}  & AP$^{50}$ & AP$^{75}$ & AP$^{M}$ & AP$^{L}$ & \textbf{mAR}  & AR$^{50}$ & AR$^{75}$ & AR$^{M}$ & AR$^{L}$ \\ \hline
		OpenPose\cite{Cao}  & 38.2          & 51.3                   & 32.5                   & 28.8                 & 46.6                 & 44.8          & 58.7                  & 41.2                   & 29.0                  & 55.6                  \\
		single\cite{Hidalgo2019} & 41.5             & 69.4                      & 29.1                      & 38.8                     & 36.4                     & 49.4             & 73.1                      & 42.2                      & 40.1                     & 48.6                     \\ \hline
		ours                  & \textbf{51.5} & \textbf{74.0}          & \textbf{46.5}          & \textbf{45.8}         & \textbf{46.9}         & \textbf{59.7} & \textbf{77.1}          & \textbf{57.4}          & \textbf{53.2}         & \textbf{56.7}         \\ \hline
	\end{tabular}
	\caption{Results of OpenPose and our method on the validation dataset.}
\end{table}
We have found that the foot and body part perform the best. Even if the human instances are small or crowded, they still works well. The face part performs well most of time, but when faces are too small or partly hidden, the detection accuracy will decrease. The hand detection, however, performs the worst and is easy to fail or make mistakes when hands are too small or occluded.

\begin{figure}[H] 
	\centering 
	\begin{minipage}[b]{0.48\textwidth} 
		\centering 
		\includegraphics[width=1\textwidth]{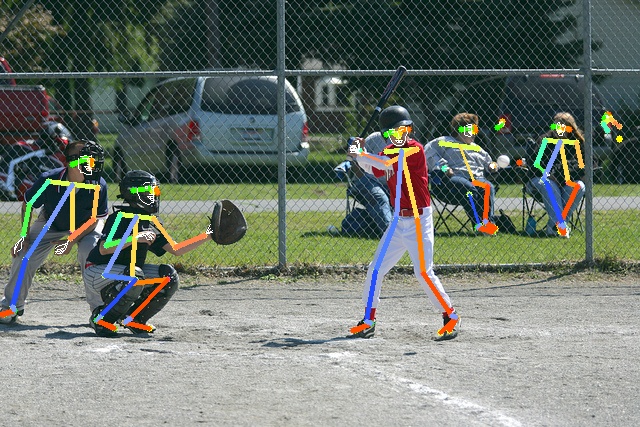} 
	\end{minipage}%
	\hspace{0.02\textwidth}%
	\begin{minipage}[b]{0.48\textwidth} 
		\centering 
		\includegraphics[width=1\textwidth]{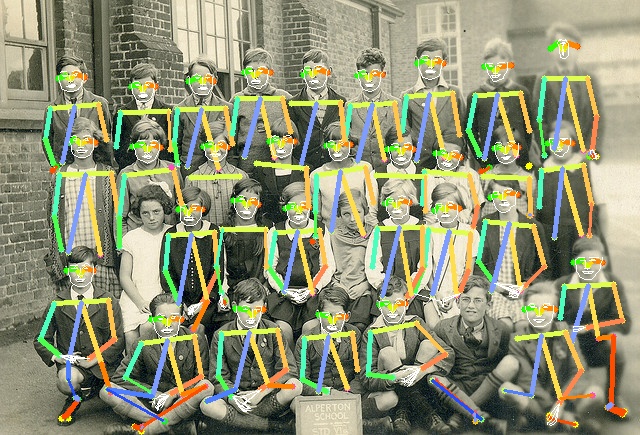} 
	\end{minipage}
	\caption{Two of the visulization results of our work.}
\end{figure}

The failures mainly result from the pseudo label, because the PRNet itself perform not well on small and occluded faces and the accuracy of hand detection in Openpose is limited. In other words, there originally exists errors in the ground truth we used for training, which is certain to influence the final result. But anyway, our method has been proved to be correct and effective. In general, our contribution is that we have proposed a novel, better method of full body pose estimation, obtained the full body pseudo labels and trained a full body pose estimation model better than previous ones.

\vspace{-1em}
\bibliographystyle{plain}
\bibliography{imvip2}
\end{document}